\documentclass[11pt,a4paper]{scrartcl}
\usepackage[round]{natbib}

\usepackage[top=3cm, bottom=3cm, left=3cm, right=3cm]{geometry}

\usepackage{booktabs}
\usepackage{color}
\usepackage{latexsym}              
\usepackage{amsmath}               
\usepackage{amssymb}               
\usepackage{amsfonts}              
\usepackage{amsthm}                
\usepackage{multirow}
\usepackage{tikz}                  
\usetikzlibrary{arrows,positioning,shapes}
\usepackage{tcolorbox}

\usepackage[english]{babel} 

\RequirePackage[%
  pdfstartview=FitH,%
  breaklinks=true,%
  bookmarks=true,%
  colorlinks=true,%
  linkcolor= blue,
  anchorcolor=blue,%
  citecolor=blue,
  filecolor=blue,%
  menucolor=blue,%
  urlcolor=blue%
  ]{hyperref}
  
  \AtBeginDocument{%
  \hypersetup{%
    pdfauthor={Benjamin Guedj},%
  	urlcolor = blue,%
  	linkcolor = blue,%
  	citecolor = orange,%
    pdftitle={Title - compilation: \today}%
  }
}

\usepackage[utf8]{inputenc} 
\usepackage[T1]{fontenc}    
\usepackage{hyperref}       
\usepackage{url}            
\usepackage{booktabs}       
\usepackage{amsfonts}       
\usepackage{nicefrac}       
\usepackage{microtype}      

\usepackage[english]{babel}
\usepackage[T1]{fontenc}
\usepackage{amsfonts}
\usepackage{dsfont}
\usepackage{amscd}
\usepackage{bbm}
\usepackage{color}
\usepackage{soul}
\usepackage{mathrsfs}
\usepackage{mathabx}
\usepackage{enumitem}
\usepackage{pgf,tikz}
\usetikzlibrary{arrows}
\usepackage{algorithm}
\usepackage{algorithmic}
\usepackage{appendix}
\usepackage{cancel}

\usepackage[capitalize]{cleveref}
\usepackage[normalem]{ulem}
\usepackage{cancel}

\usepackage{pstricks,pst-node,graphicx,graphics} 
\usepackage{caption}

\newcommand{\KL}{\operatorname{KL}}
\newcommand{\SBC}{\texttt{HYPE}}
\newcommand{\SB}{\texttt{SB}}

\allowdisplaybreaks

\usepackage[mathcal]{eucal}
\usepackage{cleveref}
\crefname{assumption}{Assumption}{Assumptions}
\crefname{equation}{Eq.}{Eqs.}
\crefname{figure}{Fig.}{Figs.}
\crefname{table}{Table}{Tables}
\crefname{section}{Sec.}{Secs.}
\crefname{theorem}{Thm.}{Thms.}
\crefname{lemma}{Lemma}{Lemmas}
\crefname{corollary}{Cor.}{Cors.}
\crefname{example}{Example}{Examples}
\crefname{appendix}{Appendix}{Appendixes}
\crefname{remark}{Remark}{Remark}

\renewenvironment{proof}[1][\proofname]{{\bfseries #1.}}{\qed \\ }

\makeatother

\theoremstyle{plain}  
\newtheorem{theorem}{Theorem}[section]
\newtheorem{definition}[theorem]{Definition}

\newtheorem{lemma}[theorem]{Lemma}
\newtheorem{proposition}[theorem]{Proposition}

\newtheorem{remark}[theorem]{Remark}

\begin{document}
\title{PAC-Bayes unleashed: generalisation bounds with unbounded losses}

\author{\textbf{Maxime Haddouche} \\ [2ex]
Inria and ENS Paris-Saclay \\
France \\\\
\textbf{Benjamin Guedj} \\ [2ex]
Inria and University College London \\
France and United Kingdom \\\\
\textbf{Omar Rivasplata} \\ [2ex]
University College London \\
United Kingdom \\\\
\textbf{John Shawe-Taylor} \\ [2ex]
University College London \\
United Kingdom \\\\
}
\date{}

\maketitle

\begin{abstract}
We present new PAC-Bayesian generalisation bounds for learning problems with unbounded loss functions. This extends the relevance and applicability of the PAC-Bayes learning framework, where most of the existing literature focuses on supervised learning problems with a bounded loss function (typically assumed to take values in the interval [0;1]). In order to relax this assumption, we propose a new notion called HYPE (standing for \emph{HYPothesis-dependent rangE}), which effectively allows the range of the loss to depend on each predictor. Based on this new notion we derive a novel PAC-Bayesian generalisation bound for unbounded loss functions, and we instantiate it on a linear regression problem. To make our theory usable by the largest audience possible, we include discussions on actual computation, practicality and limitations of our assumptions.
\end{abstract}

\section{Introduction}

Since its emergence in the late 90s, the PAC-Bayes theory (see the seminal papers by \citealp{STW1997} and \citealp{McAllester1998, McAllester1999}, or the recent survey by \citealp{guedj2019primer}) has been a powerful tool to obtain generalisation bounds and derive efficient learning algorithms. PAC-Bayes bounds were originally meant for binary classification problems \citep{seeger2002,langford2005tutorial,catoni2007} but the literature now includes many contributions involving any bounded loss function 
(without loss of generality, with values in $[0;1]$), not just the binary loss. Generalisation bounds are helpful to ensure that a learning algorithm will have a good performance on future similar batches of data.
Our goal is to provide new PAC-Bayesian generalisation bounds holding for unbounded loss functions, and thus extend the usability of PAC-Bayes to a much larger class of learning problems. 

Some ways to circumvent the bounded range assumption on the losses have been addressed in the recent literature. For instance, one approach assumes sub-gaussian or sub-exponential tails of the loss~\citep{alquier2016,germain2016}, however this requires the knowledge of additional parameters.
Some other works have also looked into the analysis for heavy-tailed losses, e.g. \cite{Alquier_2017} proposed a polynomial moment-dependent bound with $f$-divergences, while \cite{holland2019} devised an exponential bound which assumes that the second (uncentered) moment of the loss is bounded by a constant (with a truncated risk estimator, as recalled in \cref{sec:softening_functions}).
A somewhat related approach was also explored by \cite{kuzborskij2019efron}, who do not assume boundedness of the loss, but instead control higher-order moments of the generalization gap through the Efron-Stein variance proxy.

We investigate a different route here. We introduce the \emph{HYPothesis-dependent rangE} condition (\texttt{HYPE)}, which means that the loss is upper bounded by a term which does not depend on data but only on the chosen predictor for the considered learning problem. We designed this condition to be easy to verify in practice, given an explicit formulation of the loss function. Our purpose is to bring our framework to the attention of the largest machine learning community, and the \texttt{HYPE} is intended as an easy-to-use, friendly condition to yield theoretical guarantees, even for the less theoretically-oriented audience. Our regression example illustrates that purpose, and shows that a mere use of the triangle inequality is enough to check that \texttt{HYPE} is satisfied in a naive learning problem

Classical PAC-Bayes bounds  \citep[see, \emph{e.g.}][]{McAllester1999,seeger2002} have been designed with few technical conditions. For instance, besides assuming a bounded loss function with values in the interval $[0,1]$, McAllester's bound only requires absolute continuity between two densities. 
We intend to keep the same level of streamlined clarity in our assumptions, and hope practitioners could readily check whether our results apply to their particular learning problem.

\paragraph{Our contributions are twofold.}
(i) We propose PAC-Bayesian bounds holding with unbounded loss functions, therefore overcoming a limitation of the mainstream PAC-Bayesian literature for which a bounded loss is usually assumed. (ii) We analyse the bound, its implications, limitations of our assumptions, and their usability by practitioners. We hope this will extend the PAC-Bayes framework into a widely usable tool for a significantly wider range of problems, such as unbounded regression or reinforcement learning problems with unbounded rewards.

\paragraph{Outline.} 
\Cref{sec:notation} introduces our notation and definition of the \texttt{HYPE} condition.
\Cref{sec:selfb} provides a general PAC-Bayesian bound, which is valid for any learning problem complying with a mild assumption. The novelty of our approach lies in the proof technique: we adapt the notion of \emph{self-bounding function}, introduced by \cite{boucheron2000} and further developed in \cite{boucheron_concentration_inequalities2003,boucheron_concentration_inequalities2009}.
For the sake of completeness, we present in \Cref{sec:bounded_case} how our approach (designed for the unbounded case) behaves in the bounded case. This section is not the core of our work but rather serves as a safety check and particularises our bound to more classical PAC-Bayesian assumptions. 
\Cref{sec:softening_functions} introduces the notion of \emph{softening functions} and particularises \cref{sec:selfb}'s PAC-Bayesian bound. In particular, we make explicit all terms in the right-hand side.
\Cref{sec:gaussian_prior} extends our results to linear regression \citep[which has been studied from the perspective of PAC-Bayes in the literature, most recently by][]{shalaeva2019}.
Finally \Cref{sec:experiment_linear_regression} contains numerical experiments to illustrate the behaviour of our bounds in the aforementioned linear regression problem.

We defer the following material to the appendix: \Cref{sec:simus} contains additional numerical experiments for the bounded case. \Cref{sec:existing_work} presents in details related works. We reproduce in \Cref{sec:naive approach} a naive approach which inspired our study, for the sake of completeness.
\Cref{sec:corollary_softening} contains a non-trivial corollary for \cref{th: pac_bayes_bound_softening_functions}. Finally, \Cref{sec:proofs} contains all proofs to original claims we make in the paper.

\section{Notation}\label{sec:notation}

The learning problem is specified by the data space $\mathcal{Z}$, a set $\mathcal{H}$ of predictors, and a loss function $\ell : \mathcal{H}\times \mathcal{Z} \rightarrow \mathbb{R}^{+} $.
We will denote by $\mathcal{S}$ a size-$m$ dataset: $\mathcal{S}=(z_1,...,z_m) \in\mathcal{Z}^m$ where data is sampled from the same data-generating distribution $\mu$ over $\mathcal{Z}$.
For any predictor $h\in\mathcal{H}$, we define the \emph{empirical risk} $R_m(h)$ and the \emph{theoretical risk} $R(h)$ as
\[
R_m(h)= \frac{1}{m}\sum_{i=1}^m \ell(h,z_i)
\hspace{7mm}\text{and}\hspace{7mm}
R(h)= \mathbb{E}_{\mu}[\ell(h,Z)] = \mathbb{E}_{\mathcal{S}}[R_m(h)] 
\]
respectively, $\mathbb{E}_{\mu}$ denotes the expectation under $\mu$, and $\mathbb{E}_{\mathcal{S}}$ the expectation under the distribution of the $m$-sample $\mathcal{S}$. We define the generalisation gap $\Delta(h)= R(h)-R_m(h)$.
We now introduce the key concept to our analysis.
\begin{definition}[Hypothesis-dependent range (HYPE) condition]
A loss function $\ell : \mathcal{H} \times \mathcal{Z} \to \mathbb{R}^+$ 
is said to satisfy the \textbf{hypothesis-dependent range} (\texttt{HYPE}) if there exists a function $K : \mathcal{H} \to \mathbb{R}^+\backslash\{0\}$ such that $\sup_{z\in\mathcal{Z}}\ell(h,z) \leq K(h)$ for any predictor $h$.
We then say that $\ell$ is $\SBC(K)$ compliant.
\end{definition}
Let $\mathcal{M}_{1}^{+}\left(\mathcal{H}\right)$ be a set of probability distributions on $\mathcal{H}$. For $P,P' \in \mathcal{M}_{1}^{+}\left(\mathcal{H}\right)$, the notation $P' \ll P$ stands for $P'$ absolutely continuous with respect to $P$ (i.e. $P'(A) = 0$ if $P(A) = 0$ for $A$ an element of the considered $\sigma$-algebra).

We now recall a result from \citet{germain2009}. Note that while implicit in many PAC-Bayes works (including theirs), we make explicit that both the prior $P$ and the posterior $Q$ must be absolutely continuous with respect to each other. We discuss this restriction below.

\begin{theorem}[Adapted from \citealp{germain2009}, Theorem 2.1]
\label{th: germain2009}
For any $P\in\mathcal{M}_{1}^{+}(\mathcal{H})$ with no dependency on data, for any convex function $D:\mathbb{R}^{+}\times\mathbb{R}^{+}\rightarrow\mathbb{R}$, for any $\alpha\in\mathbb{R}$ and for any $\delta\in[0:1]$, we have with probability at least $1-\delta$ over size-$m$ samples $\mathcal{S}$, 
for any $Q\in\mathcal{M}_{1}^{+}(\mathcal{H})$ such that $Q \ll P$ and $P \ll Q$:
\begin{align*}
    D\left(\mathbb{E}_{h\sim Q}\left[ R_m(h)\right], \mathbb{E}_{h\sim Q}\left[ R(h)\right]\right) 
    &\leq \frac{1}{m^{\alpha}}\left(\KL(Q||P) + \log\left(\frac{1}{\delta} \mathbb{E}_{h\sim P} \mathbb{E}_{\mathcal{S}}\; e^{m^{\alpha}D(R_m(h),R(h))}  \right)\right).
\end{align*}
\end{theorem}
The proof is deferred to \cref{proof germain2009}.
Note that the proof in \cite{germain2009} does require that $P \ll Q$ although it is not explicitly stated: we highlight this in our proof. While $Q \ll P$ is classical and necessary for the $\KL(Q||P)$ to be meaningful, $P\ll Q$ appears to be more restrictive. In particular, we have to choose $Q$ such that it has the exact same support as $P$ (\emph{e.g.}, choosing a Gaussian and a truncated Gaussian is not possible).
However, we can still apply our theorem when $P$ and $Q$ belong to the same parametric family of distributions, \emph{e.g.} both `full-support' Gaussian or Laplace distributions, among others.

Note also that \citet[Theorem 4.1]{alquier2016}, which adapts a result from \citet{catoni2007}, only require $Q \ll P$. This comes at the expense of a \emph{Hoeffding's assumption} \citep[Definition 2.3]{alquier2016}. This means that $$\chi := \mathbb{E}_{h\sim P} \mathbb{E}_{\mathcal{S}}\; e^{m^{\alpha}D(R_m(h),R(h))}$$ (when $D(x,y)=x-y$ or $y-x$) is assumed to be bounded by a function only depending on hyperparameters (such as the dataset size $m$ or parameters given by Hoeffding's assumption). Our analysis does not require this assumption, which might prove restrictive in practice.

Our \cref{th: germain2009} may be seen as a basis to recover many classical PAC-Bayesian bounds. For instance, $D(x,y)=(x-y)^2$ recovers McAllester's bound \citep[as recalled in][Theorem 1]{guedj2019primer}.
To get a usable bound the outstanding task is to bound $\chi$. Note that a previous attempt has been made in \cite{germain2016}, as described in \cref{germain_2016_subsection}.

\section{Exponential moment via self-bounding functions}\label{sec:selfb}

Our goal is to control $\mathbb{E}_{\mathcal{S}}\left[ e^{m^\alpha\Delta(h)}\right]$ for a fixed $h$. 
The technique we use is based on the notion of $(a,b)$-self-bounding functions defined in \citet[Definition 2]{boucheron_concentration_inequalities2009}. 

\begin{definition}[\citealp{boucheron_concentration_inequalities2009}]
    A function $f:\mathcal{X}^m\rightarrow\mathbb{R}$ is said to be \textbf{$(a,b)$-self-bounding} with $(a,b)\in\left(\mathbb{R}^{+}\right)^2\backslash\{(0,0)\}$, if there exists $f_i :\mathcal{X}^{m-1}\rightarrow \mathbb{R}$  for every $i\in\{1..m\}$ such that $\forall i\in\{1..m\}$ and $x\in\mathcal{X}$:
    \[
    0\leq f(x) - f_i(x^{(i)}) \leq 1  
    \]
 and
 \[
 \sum_{i=1}^m f(x)-f_i(x^{(i)}) \leq af(x) +b  
 \]
 where for all $1\leq i \leq m$, the removal of the $i$th entry is $x^{(i)}= (x_1,...,x_{i-1},x_{i+1},...,x_m)$.
 We denote by $\SB(a,b)$ the class of 
 functions that satisfy this definition.
\end{definition}

In \citet[Theorem 3.1]{boucheron_concentration_inequalities2009}, the following bound has been presented to deal with the exponential moment of a self-bounding function.
Let $c_{+}:=\max(c,0)$ denote the positive part of $c \in \mathbb{R}$. We define $c_{+}^{-1}:= +\infty$ when $c_{+}=0$.

\begin{theorem}[\citealp{boucheron_concentration_inequalities2009}]
\label{th: exp_inequality2009 }
    Let $Z=g(X_1,...,X_m)$ where $X_1,...,X_m$ are independent (not necessarily identically distributed) $\mathcal{X}$-valued random variables.
    We assume that $\mathbb{E}[Z]<+\infty$. 
    If $g\in\SB(a,b)$, then defining $c=(3a-1)/6$, 
    for any $s\in[0;c_{+}^{-1})$ we have:
    \[ 
    \log\left(\mathbb{E}\left[e^{s(Z-\mathbb{E}[Z])}  \right] \right) \leq \frac{\left(a\mathbb{E}[Z] +b \right)s^2}{2(1-c_{+}s)}.   
    \]
\end{theorem}

Next, we deal with the exponential moment over $\mathcal{S}$ in \cref{th: germain2009} when $D(x,y)=y-x$. To do
so, we propose the following theorem:
\begin{theorem}
\label{th: exp_moment_bound1}
    Let $h \in \mathcal{H}$ be a fixed predictor and $\alpha\in\mathbb{R}$. 
    If the loss function $\ell$ is $\SBC(K)$ compliant, 
    then for $\Delta(h)= R(h)-R_m(h)$ we have:
\[
\mathbb{E}_{\mathcal{S}}\left[ e^{m^\alpha\Delta(h)}\right] \leq \exp\left(\frac{K(h)^2}{2m^{1-2\alpha}}\right).
\]
\end{theorem}

\begin{proof}
    We define the function $f : \mathcal{Z}^{m} \to \mathbb{R}$ as
    \[
    f:x\rightarrow \frac{1}{K(h)}\hspace{1mm}\sum_{i=1}^{m}\left(K(h)-\ell(h,x_i)\right)
    \hspace{7mm}\text{for}\hspace{2mm} x=(x_1,...,x_m)\in\mathcal{Z}^{m}.
    \]
    We also define $Z=f(Z_1,...,Z_m)$. 
    Then, notice that $\Delta(h)= \frac{K(h)}{m}\left( Z- \mathbb{E}_{\mathcal{S}}[Z] \right)$. 
    We first prove that $f\in\SB(\beta,(1-\beta)m)$ for any $\beta\in[0,1]$.
    Indeed, for all $1\leq i\leq m$, we define:
    \[
    f_i(x^{(i)})= \frac{1}{K(h)}\sum_{j\neq i} \left(K(h)-\ell(h,x_j)\right)
    \]
    where $x^{(i)}=(x_1,...,x_{i-1},x_{i+1},...,x_{m})\in\mathcal{Z}^{m-1}$ for any $x\in\mathcal{Z}^{m}$ and for any $i$.
    Then, since $0\leq \ell(h,z_i) \leq K(h)$ for all $i$, we have 
    \[
    0\leq f(z)-f_i(z^{(i)})= \frac{K(h)-\ell(h,z_i)}{K(h)} \leq 1 .
    \]
Moreover, because $f(x)\leq m$ for any $x\in\mathcal{Z}^m$, we then have:
\begin{align*}
    \sum_{i=1}^m f(x)-f_i(x^{(i)}) 
    & =  \sum_{i=1}^m \frac{K(h)-\ell(h,x_i)}{K(h)} \\
    & = f(z) =\beta f(x) + (1-\beta)f(x)
    \leq \beta f(x) + (1-\beta)m.
\end{align*}
Since this holds for any $x\in\mathcal{Z}^{m}$, this proves that $f$ is $(\beta,(1-\beta)m)$-self-bounding.

Now, to complete the proof, we will use \cref{th: exp_inequality2009 }. Because $Z$ is $(1/3,(2/3)m)$-self-bounding, we have for all $s\in\mathbb{R}^{+}$:
\[ 
\log\left(\mathbb{E}_{\mathcal{S}}\left[e^{s(Z-\mathbb{E}_{\mathcal{S}}[Z])}  \right] \right) \leq \frac{\left(\frac{1}{3}\mathbb{E}_{\mathcal{S}}[Z] +\frac{2m}{3} \right)s^2}{2}.  
\]

And since $Z\leq m$:
\begin{align*}
    \mathbb{E}_{\mathcal{S}}\left[ e^{m^\alpha\Delta(h)}\right] 
    &= \mathbb{E}_{\mathcal{S}}\left[ e^{\frac{K(h)}{m^{1-\alpha}}(Z-\mathbb{E}_{\mathcal{S}}[Z])}\right] 
        & \\
    &\leq \exp\left( \frac{\left(\frac{1}{3}\mathbb{E}_{\mathcal{S}}[Z] +\frac{2m}{3} \right)K(h)^2}{2m^{2-2\alpha}}   \right)
        & (\emph{\cref{th: exp_inequality2009 }})\\
    & \leq \exp\left( \frac{K(h)^2}{2m^{1-2\alpha}} \right).
        & (\textrm{since } \mathbb{E}_{\mathcal{S}}[Z]\leq m)
\end{align*}
\end{proof}

Comparing our \cref{th: exp_moment_bound1} with the naive result
shown in \cref{sec:naive approach} shows the strength of our approach: the trade-off lies in the fact that we are now `only' controlling $\mathbb{E}_{\mathcal{S}}\left[\exp(m^\alpha \Delta(h))\right]$ instead of $\mathbb{E}_{\mathcal{S}}\left[\exp(m^\alpha \Delta(h)^{2})\right]$, but we traded, on the right-hand side of the bound, the large exponent $m^{\alpha}K(h)^2$ for $\frac{K(h)^2}{m^{1-2\alpha}}$, the latter being much smaller when $2\alpha -1 \leq \alpha$ e.g. $\alpha\leq 1$.

Now, without any additional assumptions, the self-bounding function theory provided us a first step in our study of the exponential moment. For convenient cross-referencing, we state the following rewriting of \cref{th: germain2009}.

\begin{theorem}
\label{th: pac_bayes_with_self_bounding}
Let the loss $\ell$ being $\SBC(K)$ compliant. For any $P\in\mathcal{M}_{1}^{+}(\mathcal{H})$ with no data dependency, for any $\alpha\in\mathbb{R}$ and for any $\delta\in[0:1]$, we have with probability at least $1-\delta$ over size-$m$ samples $\mathcal{S}$, 
for any $Q$ such that $Q \ll P$ and $P \ll Q$:
\begin{align*}
    \mathbb{E}_{h\sim Q}\left[ R(h)\right] 
    \leq \mathbb{E}_{h\sim Q}\left[ R_m(h)\right] + \frac{\KL(Q||P) + \log\left(\frac{1}{\delta}\right)}{m^{\alpha}}
     +\frac{1}{m^{\alpha}}\log\left(\mathbb{E}_{h\sim P} \left[\exp\left( \frac{K(h)^2}{2m^{1-2\alpha}} \right) \right]\right).
\end{align*}
\end{theorem}

\begin{proof}
We just need to apply successively \cref{th: germain2009} with $D(x,y)=y-x$ and then \cref{th: exp_moment_bound1}.
\end{proof}

\section{Safety check: the bounded loss case} \label{sec:bounded_case}
At this stage, the reader might wonder whether this new approach allows to recover known results in the bounded case: the answer is yes.

We will, during this whole section, study the case where $\ell$ is bounded by some constant $C \in\mathbb{R}^{*}$.  We provide a bound, valid for any choice ``priors'' $P$ and ``posteriors'' $Q$ such that $Q \ll P$ and $P \ll Q$. which is an immediate corollary of \cref{th: pac_bayes_with_self_bounding}.
\begin{proposition}
    \label{p: bounded_loss_basis }
    Let $\ell$ being $\SBC(K)$ compliant, with constant $K(h)= C$, and $\alpha\in\mathbb{R}$. Then we have, for any $P\in\mathcal{M}_{1}^{+}(\mathcal{H})$ with no data dependency, 
    with probability $1-\delta$ over random $m$-samples, for any $Q\in\mathcal{M}_{1}^{+}(\mathcal{H})$ such that $Q \ll P$:
    \[
    \mathbb{E}_{h\sim Q}\left[R(h)\right] \leq \mathbb{E}_{h\sim Q}\left[R_m(h)\right] +\frac{\KL(Q||P) + \log(1/\delta)}{m^{\alpha}} + \frac{C^2}{2m^{1-\alpha}}.  
    \]
\end{proposition}

\begin{remark} We precise \cref{p: bounded_loss_basis } to evaluate the robustness of our approach, for instance, by comparing it with the PAC-Bayesian bound found in \citet{germain2016}.
This discussion can be found in \cref{germain_2016_subsection}, where the bound from \citet{germain2016} is introduced in details.
\end{remark}

\begin{remark}
    \label{rem: discussion_alpha}
    At first glance, a naive remark: in order to control the rate of convergence of all the terms of the bound in \cref{p: bounded_loss_basis } (as it is often the case in classical PAC-Bayesian bounds), then the only case of interest is in fact $\alpha=\frac{1}{2}$. However, one could notice that the factor $C^2$ is not optimisable while the KL one is. In this way, if it appears that $C^2$ is too big in practice, one wants to have the ability to attenuate its influence as much as possible and it may lead to consider $\alpha<1/2$. The following lemma is dealing with this question.
\end{remark}

\begin{lemma}
    \label{l: alpha_for_bounded_loss}
    For any given $K_{1}>0$, the function 
    $f_{K_1}(\alpha):= \frac{K_1}{m^{\alpha}} + \frac{C^{2}}{m^{1-\alpha}}$
    reaches its minimum at 
    \[
    \alpha_0= \frac{1}{2}+\frac{1}{2\log(m)}\log\left( \frac{2K_1}{C^{2}} \right).
    \]
\end{lemma}

\begin{proof}
    The explicit calculus of the $f^{'}_{K_1}$ and the resolution of $f^{'}_{K_1}(\alpha)=0$ provides the result.
\end{proof}

\begin{remark}
Our Lemma \ref{l: alpha_for_bounded_loss} indicates that if we already fixed a ``prior'' $P$ and a ``posterior'' $Q$, then taking $K_1= \KL(Q||P)+\log(1/\delta)$, offer us the optimised value of the bound given in \cref{p: bounded_loss_basis }. We numerically show in \cref{sec:simus}'s first experiment that optimising $\alpha$ leads to significantly better results .
\end{remark}

Now the only remaining question is how to optimise the KL divergence. To do so, we may need to fix an ``informed prior'' to minimise the KL divergence with an interesting posterior. This idea has been studied by \cite{lever2010, lever2013} and studied more recently by \citet{mhammedi2019,rivasplata2019}, among others. 
We will just adapt it to our problem in the most simplest way.

We will now introduce, for $k\in\{1..m\}$, the splits $\mathcal{S}_{\leq k}:= \{z_1,...,z_k\}$ and  $\mathcal{S}_{> k}:= \{z_{k+1},...,z_m\}$.

\begin{proposition}
    \label{p: data_dep_priors_prop}
    Let $\ell$ be $\SBC(K)$ compliant, with constant $K(h)= C$, and $\alpha_1,\alpha_2\in\mathbb{R}$. Then we have, for any ``priors'' $P_1\in\mathcal{M}_{1}^{+}(\mathcal{H})$ (possibly dependent on $\mathcal{S}_{>m/2}$) and $P_2\in\mathcal{M}_{1}^{+}(\mathcal{H})$ (possibly dependent on $\mathcal{S}_{\leq m/2}$), with probability $1-\delta$ over random size-$m$ samples $\mathcal{S}$,
    for any $Q\in\mathcal{M}_{1}^{+}(\mathcal{H})$ such that $Q \ll P_1$, $P_1 \ll Q$  and $Q \ll P_2$, $P_2 \ll Q$ :
    \begin{align*}
        \mathbb{E}_{h\sim Q}\left[R(h)\right] 
        & \leq \mathbb{E}_{h\sim Q}\left[R_m(h)\right] +\frac{1}{2}\left(\frac{\KL(Q||P_1) + \log(2/\delta)}{(m/2)^{\alpha_1}} + \frac{C^2}{2(m/2)^{1-\alpha_1}}\right)\\
        & + \frac{1}{2}\left(\frac{\KL(Q||P_2) + \log(2/\delta)}{(m/2)^{\alpha_2}} + \frac{C^2}{2(m/2)^{1-\alpha_2}}\right).
    \end{align*} 
\end{proposition}
\begin{proof}
    Let $P_1,P_2,Q$ as stated in the theorem. We first notice that by using \cref{p: bounded_loss_basis } on the two halves of the sample, we obtain with probability at least $1-\delta/2$:
    \[  
    \mathbb{E}_{h\sim Q}\left[R(h)\right] \leq \mathbb{E}_{h\sim Q}\left[\frac{1}{m/2}\sum_{i=1}^{m/2}\ell(h,z_i)\right] +\frac{\KL(Q||P_1) + \log(2/\delta)}{(m/2)^{\alpha_1}} + \frac{C^2}{2(m/2)^{1-\alpha_1}}  
    \]
    and also with probability at least $1-\delta/2$:
    \[  
    \mathbb{E}_{h\sim Q}\left[R(h)\right] \leq \mathbb{E}_{h\sim Q}\left[\frac{1}{m/2}\sum_{i=1}^{m/2}\ell(h,z_{m/2 +i})\right] +\frac{\KL(Q||P_2) + \log(2/\delta)}{(m/2)^{\alpha_2}} + \frac{C^2}{2(m/2)^{1-\alpha_2}}.  
    \]
    Hence with probability at least $1-\delta$ both inequalities hold, and the result follows by adding them and dividing by 2.
\end{proof}
\begin{remark}
One can notice that the main difference between \cref{p: data_dep_priors_prop} and \cref{p: bounded_loss_basis } lies in the implicit PAC-Bayesian paradigm saying that our prior must not depend on the data. With this last proposition, we implicitly allow $P_1$ to depend on $\mathcal{S}_{> m/2}$ and $P_2$ on $\mathcal{S}_{\leq m/2}$, which can in practice lead to far more accurate priors. We numerically show this fact in \cref{sec:simus}'s second experiment.
\end{remark}

\section{PAC Bayesian bounds with smoothed estimator}\label{sec:softening_functions}

We now move on to control the right-hand side term in \cref{th: pac_bayes_with_self_bounding} when $K$ is not constant. A first step is to consider a transformed estimate of the risk, inspired by the truncated estimator from \cite{catoni2012challenging}, also used in \cite{catoni2017} and more recently studied by \cite{holland2019}. The following is inspired by the results of \cite{holland2019} which we summarise in \cref{subsec:holland}.

The idea is to modify the estimator $R_m(h)$ for any $h$ by introducing a threshold $s$ and a function $\psi$ which will attenuate the influence of the empirical losses $(\ell(h,z_i))_{i=1..m}$ that exceed $s$. 
\begin{definition}[$\psi$-risks]
    For every $s>0$, $\psi: \mathbb{R}^{+} \rightarrow \mathbb{R}^{+}$, for any $h\in\mathcal{H}$, we define the \textit{empirical $\psi$-risk} $R_{m,\psi,s}$ and the \textit{theoretical $\psi$-risk} $R_{\psi,s}$  as follows:
    \[ 
    R_{m,\psi,s}(h):= \frac{s}{m}\sum_{i=1}^{m} \psi\left(\frac{\ell(h,z_i)}{s}\right) 
    \hspace{5mm}\text{and}\hspace{5mm}
    R_{\psi,s}(h):=\mathbb{E}_{\mathcal{S}}\left[R_{m,\psi,s}(h)\right] = \mathbb{E}_{\mu}\left[s\;\psi\left(\frac{\ell(h,z)}{s}\right)\right] 
    \]
    where $z\sim\mu$.
\end{definition}

We now focus on what we call \emph{softening functions}, \emph{i.e.} functions that will temperate high values of the loss function $\ell$.

\begin{definition}[Softening function]
    We say that $\psi: \mathbb{R}^{+} \rightarrow \mathbb{R}^{+}$ is a softening function if:
    \begin{itemize}
        \item $\forall x\in [0;1],\psi(x)= x$,
        \item $\psi$ is non-decreasing,
        \item $\forall x\geq 1, \psi(x)\leq x$.
    \end{itemize}
    We let $\mathcal{F}$ denote the set of all softening functions.
\end{definition}

\begin{remark}
    Notice that those three assumptions ensure that $\psi$ is continuous at $1$.
    For instance, the functions 
    $f:x\mapsto x\mathds{1}\{x\leq 1\} + \mathds{1}\{x> 1\}$ and $g:x\mapsto x\mathds{1}\{x\leq 1\} + (2\sqrt{x}-1)\mathds{1}\{x> 1\}$ are in $\mathcal{F}$.
    In \cref{subsec:holland} we compare these softening functions and those used by \citet{holland2019}.
\end{remark}
Using $\psi\in\mathcal{F}$, for a fixed threshold $s>0$, the softened loss function $s\psi\left(\frac{\ell(h,z)}{s}\right)$ verifies for any $h\in \mathcal{H},z\in \mathcal{Z}$:
\[
s\;\psi\left(\frac{\ell(h,z)}{s}\right)\leq s\;\psi\left(\frac{K(h)}{s}\right)
\]
because $\psi$ is non-decreasing. In this way, the exponential moment in \cref{th: pac_bayes_with_self_bounding} can be far more controllable. The trade-off lies in the fact that softening $\ell$ (instead of taking directly $\ell$) will deteriorate our ability to distinguish between two bad predictions when both of them are greater than $s$. 
For instance, if we choose $\psi\in\mathcal{F}$ such as $\psi = 1$ on $[1; +\infty)$ and $s>0$, if $\psi\left(\ell(h,z)/s\right)=1$ for a certain pair $(h,z)$, then we cannot tell how far $\ell(h,z)$ is from $s$ and we only can affirm that $\ell(h,z)\geq s$.

We now move on to the following lemma which controls the shortfall between $\mathbb{E}_{h \sim Q}[R(h)]$ and $\mathbb{E}_{h \sim Q}[R_{\psi,s}(h)]$ for all $Q\in\mathcal{M}_{1}^{+}(\mathcal{H})$, for a given $\psi$ and $s>0$. To do that we assume
that $K$ admits a finite moment under any posterior distribution:
\begin{align}
\label{hypothesis_for_softening_functions}
    \forall Q\in\mathcal{M}_{1}^{+}(\mathcal{H}),\hspace{2mm} \mathbb{E}_{h \sim Q}[K(h)] < +\infty.
\end{align}
For instance, if we work in $\mathcal{H}=\mathbb{R}^{N}$ and if $K$ is polynomial in $||h||$ (where $||.||$ denote the Euclidean norm), then this assumption holds if we consider Gaussian priors and posteriors.

\begin{lemma}
    \label{l: softening functions}
    Assume that \cref{hypothesis_for_softening_functions} holds, and let $\psi\in\mathcal{F}$, $Q\in\mathcal{M}_{1}^{+}(\mathcal{H}),s>0$. We have:
    \[  
    \mathbb{E}_{h\sim Q}[R(h)] \leq \mathbb{E}_{h\sim Q}[R_{\psi,s}(h)] + \mathbb{E}_{h \sim Q}\left[K(h)\mathds{1}\left\{K(h)\geq s\right\}\right].    
    \]
\end{lemma}

\begin{proof}
     Let $\psi\in\mathcal{F}$, $Q\in\mathcal{M}_{1}^{+}(\mathcal{H}),s>0$. We have, for $h\in\mathcal{H}$ :
     \begin{align*}
         R(h)- R_{\psi,s}(h) 
         & = \mathbb{E}_{z\sim\mu}\left[\ell(h,z)-s\psi\left(\frac{\ell(h,z)}{s}\right)\right] 
            & \\[1.5mm]
         & \hspace*{-20mm}= \mathbb{E}_{z\sim\mu}\left[\left(\ell(h,z)-s\psi\left(\frac{\ell(h,z)}{s}\right)\right)\mathds{1}\{\ell(h,z)\geq s\}\right] 
            & \left(\forall x\in[0,1], \psi(x)= x \right)\\
         & \hspace*{-20mm}= \mathbb{E}_{z\sim\mu}\left[\left(\ell(h,z)-s\psi\left(\frac{\ell(h,z)}{s}\right)\right)\mathds{1}\{\ell(h,z)\geq s\}\mathds{1}\left\{K(h)\geq s\right\} \right]
            & \left( \ell(h,z)\leq K(h) \right)\\
         & \hspace*{-20mm}\leq \mathbb{E}_{z\sim\mu}\left[\ell(h,z)\mathds{1}\{\ell(h,z)\geq s\} \right]\mathds{1}\left\{K(h)\geq s\right\} 
            & \left(\psi\geq 0 \right) \\
         & \hspace*{-20mm}\leq K(h)\mathbb{P}_{z\sim\mu}\left\{\ell(h,z)\geq s\right\}\mathds{1}\left\{K(h)\geq s\right\}  
            & \left( \ell(h,z)\leq K(h)\right)
     \end{align*}
     Finally, by crudely bounding the probability by 1, we get:
    \[
    R(h)\leq R_{\psi,s}(h) + K(h) \mathds{1}\left\{K(h)\geq s\right\}.
    \]
    Hence the result by integrating over $\mathcal{H}$ with respect to $Q$.
\end{proof}

Finally we present the following theorem, which provides a PAC-Bayesian inequality bounding the theoretical risk by the empirical $\psi$-risk for $\psi\in\mathcal{F}$:

\begin{theorem}
\label{th: pac_bayes_bound_softening_functions}
Let $\ell$ being $\SBC(K)$ compliant and assume that $K$ is satisfying \cref{hypothesis_for_softening_functions}. Then for any $P\in\mathcal{M}_{1}^{+}(\mathcal{H})$ with no data dependency, for any $\alpha\in\mathbb{R}$, for any $\psi\in\mathcal{F}$ and for any $\delta\in[0:1]$, we have with probability at least $1-\delta$ over size-$m$ samples $\mathcal{S}$, for any $Q$ such that $Q \ll P$ and $P \ll Q$:

\begin{align*}
    \mathbb{E}_{h\sim Q}\left[ R(h)\right] 
    &\leq \mathbb{E}_{h\sim Q}\left[ R_{m,\psi,s}(h)\right] + \mathbb{E}_{h\sim Q}\left[ K(h)\mathds{1}\{K(h)\geq s\}\right] +\frac{\KL(Q||P) + \log\left(\frac{1}{\delta}\right)}{m^{\alpha}}\\
    &\hspace{7mm} +\frac{1}{m^{\alpha}}\log\left(\mathbb{E}_{h\sim P} \left[\exp\left( \frac{s^2}{2m^{1-2\alpha}}\psi\left(\frac{K(h)}{s}\right)^2 \right) \right]\right).
\end{align*}
\end{theorem}

\begin{proof}
     Let $\psi\in\mathcal{F}$, we define the $\psi$-loss:
     \[
     \ell_2(h,z)= s\psi\left(\frac{\ell(h,z)}{s}\right) 
     \]
     Since $\psi$ is non decreasing, we have for all $(h,z)\in\mathcal{H}\times\mathcal{Z}$: 
     \[
     \ell_2(h,z)\leq s\psi\left(\frac{K(h)}{s}\right):= K_2(h)
     \]
     Thus, we apply \cref{th: pac_bayes_with_self_bounding} to the learning problem defined with $\ell_2$: for any $\alpha$ and $\delta\in(0,1)$, with probability at least $1-\delta$ over size-$m$ samples $\mathcal{S}$, for any $Q$ such that $Q \ll P$ and $P \ll Q$ we have:
\begin{align*}
    \mathbb{E}_{h\sim Q}\left[ R_{\psi,s}(h)\right] 
    &\leq \mathbb{E}_{h\sim Q}\left[ R_{m,\psi,s}(h)\right] + \frac{\KL(Q||P) + \log\left(\frac{1}{\delta}\right)}{m^{\alpha}}\\
    & +\frac{1}{m^{\alpha}}\log\left(\mathbb{E}_{h\sim P} \left[\exp\left( \frac{K_2(h)^2}{2m^{1-2\alpha}} \right) \right]\right).
\end{align*}
     We then add $\mathbb{E}_{h \sim Q}\left[K(h)\mathds{1}\left\{K(h)\geq s\right\}\right]$ on both sides of the latter inequality and apply \cref{l: softening functions}.
\end{proof}

\begin{remark}
    Notice that for every posterior $Q$, the function $\psi:x\mapsto x\mathds{1}\{x\leq 1\} + \mathds{1}\{x> 1\}$ is such that $\mathbb{E}_{h\sim P} \left[\exp\left( \frac{s^2}{2m^{1-2\alpha}}\psi\left(\frac{K(h)}{s}\right)^2 \right) \right] < +\infty$. 
    Thus, one strength of \cref{th: pac_bayes_bound_softening_functions} is to provide a PAC-Bayesian bound valid for any measure verifying \cref{hypothesis_for_softening_functions}. The choice of $\psi$ minimising the bound is still an open problem.
\end{remark}

\begin{remark}
     For the sake of clarity, we establish in \cref{sec:corollary_softening} a corollary of \cref{th: pac_bayes_bound_softening_functions} (with an assumption which is stronger than \cref{hypothesis_for_softening_functions}) which is easier to compare to the result of \cite{holland2019}.
 \end{remark}

\section{The linear regression problem}\label{sec:gaussian_prior}

We now focus on the celebrated linear regression problem and see how our theory translates to that particular learning problem. We assume that data is a size-$m$ sample $(z_i)_{i=1..m}$ drawn independently under the distribution $\mu$, where for all $i$, $z_i=(x_i,y_i)$ with $x_i\in\mathbb{R}^N,y_i\in \mathbb{R}$.

Our goal here is to find the most accurate predictor $h\in\mathbb{R}^N$ with respect to the loss function $\ell(h,z)= |\langle h,x \rangle -y|$, where $z=(x,y)$. 
We will make the following mild assumption: there exists $B,C \in \mathbb{R}{+}\backslash\{0\}$ such that for all $z=(x,y)$ drawn under $\mu$:
\[||x|| \leq B \;\text{ and }\; |y| \leq C\]
where $||.||$ is the norm associated to the classical inner product of $\mathbb{R}^{N}$.
Under this assumption we note that for all $z=(x,y)$ drawn according to $\mu$, we have:
\begin{align*}
    \ell(h,z) 
    = |\langle h,x \rangle -y| \leq  |\langle h,x \rangle| +|y] 
    \leq ||h||.||x|| + |y|
    \leq B||h|| + C.
\end{align*}
Thus we define $K(h)= B||h|| +C$ for $h\in\mathbb{R}^{N}$.
If we first restrict ourselves to the framework of \cref{sec:selfb}, we want to use \cref{th: pac_bayes_with_self_bounding} and doing so, our goal is to bound $\xi := \mathbb{E}_{h\sim P} \left[\exp\left( \frac{K(h)^2}{2m^{1-2\alpha}} \right) \right]$. 
The shape of $K$ invites us to consider a Gaussian prior. Indeed, we notice that if  $P=\mathcal{N}(0,\sigma^{2}\textbf{I}_{N})$ with $0< \sigma^{2}< \frac{m^{1-2\alpha}}{B^2} $, then $\xi < +\infty$.
Notice that we cannot take just any Gaussian prior, however with a small $\alpha$, the condition $0< \sigma^{2}< \frac{m^{1-2\alpha}}{B^2} $ may become quite loose. Thus, we have the following: 
\begin{theorem}
\label{th: pacbayes_bound_gaussian_case}
\noindent
 Let $\alpha\in\mathbb{R}$ and $N\geq 6$. If the loss $\ell$ is $\SBC(K)$ compliant with $K(h)= B||h|| + C $, with $B>0,C\geq 0$, then we have, for any Gaussian prior $P=\mathcal{N}(0,\sigma^{2}\textbf{I}_{N})$ with $ \sigma^{2}= t\frac{m^{1-2\alpha}}{B^2}$, \hspace{1mm} $0<t<1$. 
 We have with probability $1-\delta$ over size-$m$ samples $\mathcal{S}$,
 for any $Q\in\mathcal{M}_{1}^{+}(\mathcal{H})$ such that $Q \ll P$ and $P \ll Q$:
 \begin{align*}
     \mathbb{E}_{h\sim Q}[ R(h)] 
     & \leq \mathbb{E}_{h\sim Q}[R_m(h)] + \frac{\KL(Q||P)+ \log\left(2/\delta\right)}{m^{\alpha}} + \frac{C^2}{2m^{1-\alpha}}\left(1+ f(t)^{-1}\right)   \\
     & + \frac{N}{m^{\alpha}}\left( \log\left( 1+ \left(\frac{C}{\sqrt{2f(t)m^{1-2\alpha}}}\right)\right) + \log\left(\frac{1}{\sqrt{1-t}}\right)  \right)
 \end{align*}
where $f(t)=\frac{1-t}{t}$.
\end{theorem}

The proof is deferred to \cref{proof_gaussian_prior}.
To compare our result with those found in the literature, we can fix $\alpha= 1/2$. Doing so, we lose the dependency in $m$ for the choice of the variance of the prior (which now only depends on $B$), but we recover the classic decreasing factor $1/\sqrt{m}$.

\begin{remark}
    Notice that for now we did not use \cref{sec:softening_functions} even if we could (because $K$ is polynomial in $||h||$ and we consider Gaussian priors and posteriors, so \cref{hypothesis_for_softening_functions} is satisfied). Doing so, we obtained a bound which appears to depend linearly on the dimension $N$. In practice $N$ may be too big, and in this case, introducing an adapted softening function $\psi$ (one can think for instance of $\psi(x)= x\mathds{1}\{x\leq 1\} + \mathds{1}\{x> 1\}$) is a powerful tool to attenuate the weight of the exponential moment. This also extends the class of authorised Gaussian priors by avoiding to stick with a variance $\sigma^{2}= t\frac{m^{1-2\alpha}}{B^2}$,\hspace{1mm} $0<t<1$.
\end{remark}

\section{Numerical experiments for linear regression}\label{sec:experiment_linear_regression}

\paragraph{Setting.} In this section we apply \cref{th: pacbayes_bound_gaussian_case} on a concrete linear regression problem. The situation is as follows: we want to approximate the function $f(x)=\sqrt{\langle h^*,x\rangle}$ where $h^{*}\in\mathbb{R}^d$. We assume that $h^*$ lies in an hypercube centered in $0$ of half-side $c$, \emph{e.g.} the set $\{(h_i)_{i=1,...,d}\mid \forall i, |h_i|\leq c\}$. Doing so we have $||h^*||\leq c\sqrt{d}$.

Furthermore,  we assume that data is drawn inside an hypercube of half-side $e$. Doing so we have for any data $x, ||x||\leq e\sqrt{d}$.

For any data $x$, we define $y=f(x)$ and we set $\mathcal{H}=\mathbb{R}^d$. As described in \cref{sec:gaussian_prior}, we set $\ell(h,x,y)= |\langle h,x\rangle -y|$. We then remark that for any $(h,x,y)$:
\begin{align*}
    \ell(h,x,y)& \leq |\langle h,x\rangle| +  |y| \leq ||h||||x|| + |\sqrt{\langle h^*,x\rangle}| \\
    & \leq e\sqrt{d}||h|| + \sqrt{||h^*||.||x||} \leq e\sqrt{d}||h|| + \sqrt{c\sqrt{d}.e\sqrt{d}}\\
    & \leq e\sqrt{d}||h|| + \sqrt{cde}
\end{align*}
Then we can define $B= e\sqrt{d}$ and $C=\sqrt{cde}$ to apply \cref{th: pacbayes_bound_gaussian_case}. 
We also define $\mathcal{M}_1^{+}(\mathcal{H}):=\left\{ \mathcal{N}(h,\sigma^2 I_d)\mid h\in \mathcal{H}, \sigma^2\in\mathbb{R}^+  \right\} $
which is the set of candidate measures for this learning problem. Recall that in practice, given a fixed $\alpha\in\mathbb{R}$, we are only allowed to consider priors such that their variance $\sigma^2\in \left]0;\frac{m^{1-2\alpha}}{B^2}\right[$.
 We want to learn an optimised predictor given a dataset $\mathcal{S}=((x_i,y_i))_{i=1,\dots,m}$. To do so we consider synthetic data.
\paragraph{Synthetic data.}
We draw $h^*$ under a Gaussian (with mean 0 and standard deviation equal to $5$) truncated to the hypercube centered in $0$ of half-side $c$ .
We generate synthetic data according to the following process: for a fixed sample size $m$, we draw $x_1,...,x_m$ under a Gaussian (with mean 0 and standard deviation equal to $5$) truncated to the hypercube centered in $0$ of half-side $e$.
 
\begin{figure}[H]
    \centering
    \includegraphics[scale= 0.7]{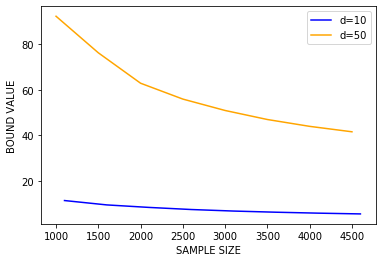}
    \caption{Evaluation of the right hand side in \cref{th: pacbayes_bound_gaussian_case} with $d=10$ and $d=50$.} 
    \label{main_experiments}
\end{figure}

\paragraph{Experiment.} First, we fix $c=e=10$. Our goal here is to obtain a generalisation bound on our problem. 
We fix arbitrarily, for a fixed $\alpha\in\mathbb{R}$, $t_0=1/2$ and $\sigma_0^2=t_0\frac{m^{1-2\alpha}}{B^2}$ and we define our \textit{naive prior} as $P_0=\mathcal{N}(0,\sigma_0^2 I_d)$. 
For a fixed dataset $\mathcal{S}$, we define our posterior as $Q(\mathcal{S}):=\mathcal{N}(\hat{h}(\mathcal{S}),\sigma^2 I_d)$, with $\sigma^2\in\{\sigma_0^2/2,...,\sigma_0^2/2^J\}, (J=\log_2(m))$ such that it is minimising the bound among candidates. 
Note that all the previously defined parameters are depending on $\alpha$, which is why we choose $\alpha\in\{i/\texttt{step}\mid 0\leq i \leq \texttt{step}\}$ for \texttt{step} a fixed integer (in practice \texttt{step}=8 or 16) and we take the value of $\alpha$ minimising the bound among the candidates as well.
 \cref{main_experiments} contains two figures, one with $d=10$, the other with $d=50$. On each figure are computed the right-hand side term in \cref{th: pacbayes_bound_gaussian_case} with an optimised $\alpha$ for each step.
 
\paragraph{Discussion.} To the the best of our knowledge, this is the first attempt to numerically compute PAC-Bayes bounds for unbounded problems, making it impossible to compare to other results. We stress though that obtaining numerical values for the bound without assuming a bounded loss is a significant first step. Furthermore, we consider a rather hard problem: $f$ is not linear, so we cannot rely on a linear approximation fitting perfectly data, and the larger the dimension, the larger the error, as illustrated by \cref{main_experiments}. Thus for any posterior $Q$, the quantity $\mathbb{E}_{h\sim Q}[ R(h)]$ is potentially large in practice and our bound might not be tight. Finally, notice that optimising $\alpha$ (instead of taking $\alpha=1/2$ to recover a classic convergence rate) leads to a significantly better bound. A numerical example of this assertion is presented in \cref{sec:simus}. We aim to conduct further studies to consider the convergence rate as an hyperparameter to optimise, rather than selecting the same rate for all terms in the bound.

\section{Conclusion}
The main goal of this paper is to expand the PAC-Bayesian theory to learning problems with unbounded losses, under the \texttt{HYPE} condition. We plan next to particularise our general theorems to more specific situations, starting with the kernel PCA setting.




\bibliography{biblio}


\appendix

\section{Additional experiments for the bounded loss case}\label{sec:simus}

Our experimental framework has been inspired of the work of \citet{mhammedi2019}.\vspace{3mm}\\
\textbf{Settings} We generate synthetic data for classification, and we are using the 0-1 loss. Here, the data space is $\mathcal{Z}=\mathcal{X}\times\mathcal{Y}= \mathbb{R}^d\times \{0,1\}$ with $d\in\mathbb{N}$. Here the set of predictors $\mathcal{H}$ is also $\mathbb{R}^d$. And for $z=(x,y)\in\mathcal{Z}, h\in\mathcal{H}$, we define the loss as $\ell(h,z):= |\mathds{1}\left\{\phi(h^{\top}x)>1/2\right\}-y|$. where $\phi(w)=\frac{1}{1+e
^{-w}}$
We want to learn an optimised predictor given a dataset $\mathcal{S}=(z_i=(x_i,y_i))_{i=1..m}$. To do so we use \textit{regularised logistic regression} and we compute:
\begin{align}
    \label{reg_log_regression}
    \hat{h}(\mathcal{S}) := \text{arg}\min_{h\in\mathcal{H}} \lambda \frac{||h||^2}{2} - \frac{1}{m}\sum_{i=1}^m y_i \log\left( \phi(h^{\top}x_i) \right) + (1-y_i)\log\left(1- \phi(h^{\top}x_i)  \right) 
\end{align}
where $\lambda$ is a fixed regularisation parameter. We also define \[\mathcal{M}_1^{+}(\mathcal{H}):=\left\{\mathcal{N}(h,\sigma^2 I_d)\mid h\in \mathcal{H}, \sigma^2\in\mathbb{R}^+   \right\} \]
which is the set of considered measures for this learning problem.\vspace{3mm}\\
\textbf{Parameters} We set $\delta= 0.05, \lambda=0.01$. We approximately solve \cref{reg_log_regression} by using the \texttt{minimize} function of the optimisation module in Python, with the Powell method. To approximate gaussian expectations, we use Monte-Carlo sampling.  \vspace{3mm}\\
\textbf{Synthetic data}
We generate synthetic data for $d=10$ according to the following process: for a fixed sample size $m$, we draw $x_1,...,x_m$ under the multivariate gaussian distribution $\mathcal{N}(0,I_d)$ and we compute for all $i$: $y_i= \mathds{1}\{\phi(h^{*\top}x_i)>1/2\}$ where $h^*$ is the vector formed by the $d$ first digits of $\pi$.\vspace{3mm}\\
\textbf{Normalisation trick}
Given the predictors shape, we notice that for any $h\in\mathcal{H}$:
\[\mathds{1}\{\phi(h^{*\top}x)>1/2\} = 1 \Leftrightarrow  \frac{1}{1 +\exp(-h^{\top}x)} >\frac{1}{2} \Leftrightarrow  h^{\top}x<0   \]
Thus, the value of the prediction is exclusively determined by the sign of the inner product, and this quantity is definitely not influenced by the norm of the vector. \\
Then, for any sample $\mathcal{S}$, we call \textbf{normalisation trick} the fact of considering $\hat{h}(\mathcal{S})/||\hat{h}(\mathcal{S})||$ instead of $\hat{h}(\mathcal{S})$ in our calculations. This process will not deteriorate the quality of the prediction and will considerably enhance the value of the KL divergence. \vspace{3mm}\\
\textbf{First experiment}
Our goal here is to highlight the point discussed in \cref{rem: discussion_alpha} e.g. the influence of the parameter $\alpha$ in \cref{p: bounded_loss_basis }. 
We fix arbitrarily $\sigma_0^2=1/2$ and we define our \textit{naive prior} as $P_0=\mathcal{N}(0,\sigma_0^2 I_d)$.
For a fixed dataset $\mathcal{S}$, we define our posterior as $P(\mathcal{S}):=\mathcal{N}(\hat{h}(\mathcal{S}),\sigma^2 I_d)$, with $\sigma^2\in\{1/2,...,1/2^J\}, (J=\log_2(m))$ such that it is minimising the bound among candidates. \\
 We computed two curves: first,  \cref{p: bounded_loss_basis } with $\alpha=1/2$ second, \cref{p: bounded_loss_basis } again with $\alpha$ equals to the value proposed in \cref{l: alpha_for_bounded_loss}. Notice that to compute this last bound, we first optimised our choice of posterior with $\alpha=1/2$ and we then optimised $\alpha$. We did this to be consistent with Lemma \ref{l: alpha_for_bounded_loss}. Indeed, we proved this lemma by assuming that the KL divergence was already fixed, hence our optimisation process in two steps. \textbf{We chose to apply the normalisation trick here}, we then obtained the left curve of \cref{experiments_bounded_case}.
 
\textbf{Discussion} From this curve, we formulate several remarks. 
First, we remark on this specifc case, our theorem provide a quite tight result in practice ( with an error rate lesser than $10\%$ for the bound with optimised alpha).\\
Secondly we can now confirm that choosing an optimised $\alpha$ leads to a tighter bound: in further studies, it will relevant to adjust $\alpha$ with regards to the different terms of our bound instead of looking for an identical convergence rate for all the terms.

\vspace{3mm}
\textbf{Second experiment}
We want now to study \cref{p: data_dep_priors_prop}  e.g. to see if an informed prior provide effectively a tighter bound than a naive one. We will use the notations introduced in \cref{p: data_dep_priors_prop}. For a dataset $\mathcal{S}$ we define $h_1(\mathcal{S})= h(\mathcal{S}_{> m/2}) $ the vector resulting of the optimisation of \cref{reg_log_regression} on  $\mathcal{S}_{> m/2}$. We define similarly $h_2(\mathcal{S}):= h(\mathcal{S}_{\leq m/2})$.
We fix arbitrarily $\sigma_0^2=1/2$ and we define our \textit{informed priors} as $P_1=\mathcal{N}(h_1(\mathcal{S}),\sigma_0^2 I_d)$ and $P_2=\mathcal{N}(h_2(\mathcal{S}),\sigma_0^2 I_d)$. Finally, we define our posterior as $P(\mathcal{S}):=\mathcal{N}(\hat{h}(\mathcal{S}),\sigma^2 I_d)$, with $\sigma^2\in\{1/2,...,1/2^J\}, (J=\log_2(m))$ with $\sigma^2$ optimising the bound among the same candidate than the first experiment. \\
 We computed two curves: first, \cref{p: bounded_loss_basis } with $\alpha$ optimised accordingly to \cref{l: alpha_for_bounded_loss} secondly, \cref{p: data_dep_priors_prop} with $\alpha_1,\alpha_2$ optimised as well and the informed priors as defined above. \textbf{We chose to not apply the normalisation trick here}, we then obtained the right curve of \cref{experiments_bounded_case}:

\begin{figure}[H]
    \centering
    \includegraphics[scale= 0.5]{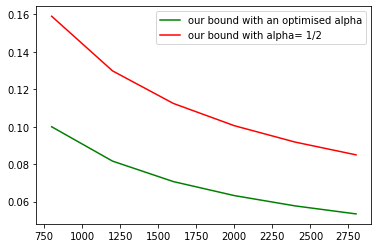}
    \includegraphics[scale= 0.5]{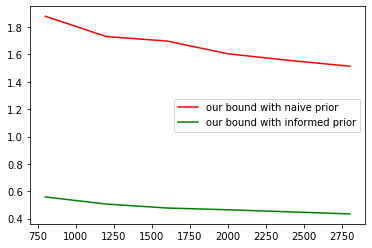}
    \caption{On the left, result of the first experiment which highlight the importance of optimising $\alpha$. On the right, result of the second experiment which show how effective an informed prior is.}
    \label{experiments_bounded_case}
\end{figure}

\paragraph{Discussion}
We clearly see that with this framework having an informed prior is a powerful tool to enhance the quality of our bound.
Notice that we voluntarily chose to not apply the normalisation trick here. The reason behind this is that this trick appears to be too powerful in practice, and applying it leads to be counterproductive to highlight our point: the bound without informed prior would be tighter than the one with. Furthermore, this trick is very linked to the specific structure of our problem and is not valid for any classification problem. Thus, the idea of providing informed priors remains an interesting tool for most of the cases.

\section{Existing work}
\label{sec:existing_work}
\subsection{Germain et al., 2016}
\label{germain_2016_subsection}

In \citet[Section 4]{germain2016}, a PAC-Bayesian bound has been provided for all \textit{sub-gamma} losses with a variance $s^2$ and scale parameter $c$, under a data distribution $\mu$ and a prior $P$, i.e. losses satisfying the following property: 
\[
\forall \lambda\in (0,\frac{1}{c}), \log\left(\frac{1}{\delta} \mathbb{E}_{h\sim P} \mathbb{E}_{\mathcal{S}}\; e^{\lambda(R(h)-R_m(h))}  \right) \leq \frac{s^2}{c^2}(-\log(1-c\lambda) - \lambda c) \leq \frac{\lambda^2s^2}{2(1-c \lambda)}. 
\]
Note that a sub-gamma loss (with regards to $\mu$ and $P$) is potentially unbounded. 
Germain et al. then propose the following PAC-Bayesian bound:

\begin{theorem}[\citealp{germain2016}]
\label{th: germain2016}
If the loss $\ell$ is sub-gamma with a variance $s^2$ and scale parameter $c$, under the data distribution $\mu$ and a fixed prior $P\in\mathcal{H}$, then we have, with probability $1-\delta$ over size-$m$ samples,
for any $Q \ll P$:
    \[
    \mathbb{E}_{h\sim Q}\left[R(h)\right] \leq \mathbb{E}_{h\sim Q}\left[R_m(h)\right] +\frac{\KL(Q||P) + \log(1/\delta)}{m} + \frac{s^2}{2(1-c)}  
    \]
\end{theorem}

\cref{th: germain2016} will be quoted several times in this paper given that it is a concrete PAC Bayesian bound provided with the will to overcome the constraint of a bounded loss. It is also one of the only one found in literature by the authors.

\paragraph{Can we apply this theorem to the bounded case?} The answer is yes: we remark that thanks to Hoeffding's lemma, if $\ell$ is bounded by $C$, then for any $h\in\mathcal{H}$, $R_m(h)-R(h)\in[-C,C]$ almost surely. So,  $\forall \lambda\in\mathbb{R},$ $\log \mathbb{E}_{z\sim\mu}\left[e^{\lambda(R(h)-R_m(h)}\right]\leq \frac{\lambda^{2}C^2}{2} $. So, for any prior $P$, $\log \mathbb{E}_{h\sim P} \mathbb{E}_{z\sim\mu}\left[e^{\lambda(R(h)-R_m(h)}\right]\leq \frac{\lambda^{2}C^2}{2} $.
\\
Thus, $\ell$ is sub-gamma with variance $C^2$ and scale parameter 0. 
So \cref{th: germain2016} can be applied with $s^2= C^2$, $c=0$.

\paragraph{Comparison with \cref{p: bounded_loss_basis }}  We remark that by taking $K=C$ and $\alpha=1$ in \cref{p: bounded_loss_basis }, we are recovering \cref{th: germain2016}. However, our approach allows us to say that if we can obtain a more precise form of $K$ such that $\forall h\in\mathcal{H}$, $K(h)\leq C$ and $K$ is non-constant, \cref{th: pac_bayes_with_self_bounding}, will ensure us that 
\[ 
\frac{1}{m^{\alpha}}\log\left(\mathbb{E}_{h\sim P} \left[\exp\left( \frac{K(h)^2}{2m^{1-2\alpha}} \right) \right]\right)\leq \frac{C^2}{2m^{1-\alpha}} 
\]
Thus, having a precise information on the behavior of the loss function $\ell$ with regards to the predictor $h$ allows us to obtain a tighter control of the exponential moment, hence a tighter bound.

\begin{remark}
    We can see that \cref{th: germain2016}, one can't control the factor $C^2/2$. However, the authors remarked this apparent weakness and partially corrected this issue on \cite[Section 4, Eq (13),(14)]{germain2016}. Indeed, they proposed to balance the influence of $m$ between the different terms of the PAC Bayesian bound by providing to all terms the same convergence rate in $1/\sqrt{m}$.
    
    We then can see \cref{p: bounded_loss_basis } as a proper generalisation of \cite[Section 4, Eq (13),(14)]{germain2016}. Indeed, our bound exhibits properly the influence of the parameter $\alpha$. Thus, we understand (and Lemma \ref{l: alpha_for_bounded_loss} proves it) that the choice of $\alpha$ deserves a study in itself in the way it is now a parameter of our optimisation problem. This fact has already be higlightened in \citet[Theorem 4.1]{alquier2016} (where $\lambda:= m^{\alpha}$).
\end{remark}

\subsection{Holland, 2019}
\label{subsec:holland}

\cite{holland2019} proposed a PAC Bayesian inequality with unbounded loss. For that he introduced a function $\psi$ verifying  a few specific conditions, different of those we used in \cref{sec:softening_functions} to define our set of softening functions. Indeed he considered a function $\psi$ such that:

\begin{itemize}
    \item $\psi$ is bounded,
    \item $\psi$ is non decreasing,
    \item it exists $b>0$ such that for all $u\in\mathbb{R}$:
    \begin{align}
    \label{hollands_condition}
        -\log\left(1-u+\frac{u^2}{b}\right)\leq \psi(u)\leq \log\left(1+u+\frac{u^2}{b} \right).    
    \end{align}
\end{itemize}

We remark that, as Holland did, we supposed that our softening functions are non-decreasing. We chose softening functions to be equal to $Id$ on $[0,1]$ which is quite restrictive but we are just imposing softening functions to be lesser than $Id$ on $\left[1,+\infty\right)$ where Holland supposed $\psi$ to be bounded and satisfy \cref{hollands_condition}. A concrete example of such a function $\psi$ lies in the piecewise polynomial function
of \citet{catoni2017}, defined by:
\[ 
\psi(u)= \begin{cases}
-2\sqrt{2}/3 &\text{ if }u\leq -\sqrt{2} \\
u-u^3/6 &\text{ if } u\in [-2\sqrt{2}/3,2\sqrt{2}/3] \\
2\sqrt{2}/3& \text{otherwise.}
\end{cases} 
\]

As in \cref{sec:softening_functions}, we are considering the $\psi$-empirical risk $R_m,\psi,s$ for any $s>0$. Holland provided his theorem given the fact the following assumptions are realised:
\begin{itemize}
    \item Bounds on lower-order moments. For all $h\in\mathcal{H}$, we have $\mathbb{E}_{z\sim\mu}[\ell(h,z)^2]\leq M_2<+\infty$, $\mathbb{E}_{z\sim\mu}[\ell(h,z)^3]\leq M_3<+\infty$. 
    \item Bounds on the risk. For all $h\in\mathcal{H}$, we suppose $R(h)\leq \sqrt{mM_2/(4\log(\delta^{-1})}$.
    \item Large enough confidence, we require $\delta\leq e^{-1/9}$.
\end{itemize}
Now we can state Holland's theorem.
\begin{theorem}[\citealp{holland2019}]
\label{hollands_theorem}
    Let $P$ be a prior distribution on model $\mathcal{H}$. Let the three assumptions listed above hold. Setting $s^2= m M_2/(2\log(\delta^{-1}))$, then with probability at least $1 - \delta$ over the random draw of the size-$m$ sample, it holds that
\begin{align*}
 \mathbb{E}_{h\sim Q}\left[ R(h)\right] 
 & \leq \mathbb{E}_{h\sim Q}\left[ R_{m,\psi,s}(h)\right] +\frac{1}{\sqrt{m}}\left( \KL(Q||P) + \frac{1}{2}\log\left(\frac{8\pi M_2}{\delta^2}  \right) -1  \right) \\
 & + \frac{1}{\sqrt{m}} \nu^{*}(\mathcal{H}) +O\left(\frac{1}{m}\right)
 \end{align*}
 where $\nu^{*}(\mathcal{H}):=\mathbb{E}_{h\sim P} \left[\exp\left(\sqrt{m}(R(h)-R_{m,\psi,s}(h))\right)  \right]\hspace{1mm}/\hspace{1mm}\mathbb{E}_{h\sim P} \left[\exp\left(R(h)-R_{m,\psi,s}(h)\right)\right]$ 
\end{theorem}

\section{Exponential moment via tail integrals} 
\label{sec:naive approach}
This section provides a bound of the exponential moment when $D(x,y)=(x-y)^2$ by only using classic properties, i.e. without the self-bounding property.

\begin{theorem}
   Let $h$ be a fixed predictor, $\alpha\in\mathbb{R}$ and $\mathcal{S}=(z_1,...,z_m)$ be the $m$-sample of data. If the loss $\ell$ satisfies 
   the \texttt{HYPE} condition with $K(h)$, then we have:
   \[
   \mathbb{E}_{\mathcal{S}}\left[ e^{m^\alpha\Delta(h)^2}\right] \leq 1 + \frac{2}{1-\frac{m^{1-\alpha}}{2K(h)^2}} \left[ \exp\left(m^{\alpha} K(h)^2 - \frac{m}{2}\right) -1\right]. 
   \]
   Recall that $\Delta(h):= R(h)-R_m(h)$.
\end{theorem}

\begin{proof}
First let us notice that almost surely we have:
\begin{align}
    \label{useful_b}
    \forall i, \hspace{3mm}
    0\leq \ell(h,z_i)\leq K(h), \; \text{ so } \Delta(h)\leq K(h).
\end{align}
Then
\begin{align*}
    \mathbb{E}_{\mathcal{S}}\left[ e^{m^\alpha\Delta(h)^2}\right] 
    & = \int_0^{+\infty} \mathbb{P}\left( e^{m^\alpha\Delta(h)^2} >t\right)dt \\
    & \leq 1 + \int_1^{+\infty} \mathbb{P}\left(\exp\left(m^\alpha\Delta(h)^2\right) >t\right)dt  \\
    & \leq 1 + \int_0^{+\infty} \mathbb{P}\left(\exp\left(m^\alpha\Delta(h)^2\right) >e^{u^2}\right)2u e^{u^{2}}du & (t= e^{u^{2}})\\
    &\leq 1 + \int_0^{\sqrt{m}^{\alpha} K(h)} \mathbb{P}\left(\exp\left(m^\alpha\Delta(h)^2\right) >e^{u^2}\right)2u e^{u^{2}}du & \text{(thanks to \cref{useful_b})} \\
    & \leq 1 + \int_0^{\sqrt{m}^{\alpha} K(h)} \mathbb{P}\left(|\Delta(h)| >m^{\frac{\alpha}{2}}u\right)2u e^{u^{2}}du. &
\end{align*}
Thanks to \cref{useful_b}, we can use Hoeffding's inequality on $R_m(h)$. We thus obtain:
\begin{align*}
 \forall u>0, \hspace{3mm} 
 \mathbb{P}\left(|\Delta(h)| > m^{\frac{\alpha}{2}}u\right) 
 & \leq 2 \exp\left(-\frac{u^2}{2m^{\alpha} \sum_{i=1}^m (\frac{K(h)}{m})^2 }\right)\\
 & \leq 2 \exp\left(-\frac{m^{1-\alpha}u^2}{2K(h)^2}\right).
\end{align*}
So by application of this inequality and the change of variable $v=u^2$ we have:
\begin{align*}
   \mathbb{E}_{\mathcal{S}}\left[ e^{m^\alpha\Delta(h)^2}\right] 
   &\leq 1 + \int_0^{m^{\alpha} K(h)^2}
        2\exp\left(v-\frac{vm^{1-\alpha}}{2K(h)^2}\right)dv  
   \\[1mm]
   &\leq 1 + \frac{2}{1-\frac{m^{1-\alpha}}{2K(h)^2}} 
        \left[ \exp\left(m^{\alpha} K(h)^2 - \frac{m}{2}\right) -1\right]
\end{align*}
which completes the proof.
\end{proof}   

\section{A corollary of Theorem
\ref{th: pac_bayes_bound_softening_functions}} 
\label{sec:corollary_softening}
We are now dealing with the following assumption on $K$: it exists a constant $M_3$ such that:
\begin{align}
\label{strong_hypothesis_for_softening_functions}
    \sup_{Q\in\mathcal{M}_{1}^{+}(\mathcal{H})}\hspace{2mm} \mathbb{E}_{h \sim Q}[K(h)^{3}]\leq M_3 < +\infty.
\end{align}
In other words, we assume that the third moments under any posterior distribution are uniformly bounded by a fixed constant $M_3$. Thus, this is a stronger assumption than \cref{hypothesis_for_softening_functions}.\\
Under this assumption, we can properly define the (finite) following quantity:
\[ 
\forall s>0,\hspace{1mm} M_{3,s}:=  \sup_{Q\in\mathcal{M}_{1}^{+}(\mathcal{H})} \mathbb{E}_{h \sim Q}\left[K(h)^{3}\mathds{1}\left\{K(h)\geq s\right\}\right] \leq M_3.   
\]

\begin{lemma}
    \label{l: softening functions_with strong hypothesis}
    Assume that \cref{strong_hypothesis_for_softening_functions} holds and let $\psi\in\mathcal{F}$, $Q\in\mathcal{M}_{1}^{+}(\mathcal{H}),s>0$. We have :
    \[  
    \mathbb{E}_{h\sim Q}[R(h)] \leq \mathbb{E}_{h\sim Q}[R_{\psi,s}(h)] + \frac{M_{3,s}}{s^2}.    
    \]
\end{lemma}

\begin{proof}
    The beginning of the proof of \cref{l: softening functions} holds here. We then have for any $h\in\mathcal{H}$: 
    \[
    R(h)- R_{\psi,s}(h)  \leq K(h)\mathbb{P}_{z\sim\mu}\left\{\ell(h,z)\geq s\right\}\mathds{1}\left\{K(h)\geq s\right\}.   
    \]  
    Yet, by Markov's inequality, we have:
     \begin{align*}
          \mathbb{P}_{z\sim\mu}\left\{\ell(h,z) \geq s\right\} 
          & \leq \frac{\mathbb{E}_{z\sim\mu}\left[\ell(h,z)^2\right]}{s^2}  \leq \frac{K(h)^2}{s^2}.
     \end{align*}
    So we can finally affirm that: 
    \[
    R(h)\leq R_{\psi,s}(h) + \frac{K(h)^{3}}{s^2} \mathds{1}\left\{K(h)\geq s\right\}.
    \]
    Hence the result by integrating over $\mathcal{H}$ with $Q$ and bounding $\mathbb{E}_{h\sim Q}[K(h)^{3} \mathds{1}\left\{K(h)\geq s\right\}]$ by $M_{3,s}$.
\end{proof}

Finally we present the following theorem, which is a corollary of \cref{th: pac_bayes_bound_softening_functions}:

\begin{theorem}
\label{th: pac_bayes_bound_softening_functions_strong_hypothesis}
Let $\ell$ being $\SBC(K)$ compliant and assume that $K$ is satisfying \cref{strong_hypothesis_for_softening_functions}. Then for any prior $P\in\mathcal{M}_{1}^{+}(\mathcal{H})$ with no data dependency, for any $\alpha\in\mathbb{R}$, for any $\psi\in\mathcal{F}$ and for any $\delta\in[0:1]$, we have with probability at least $1-\delta$ over size-$m$ samples $\mathcal{S}$, for any $Q$ such that $Q \ll P$ and $P \ll Q$:
\begin{align*}
    \mathbb{E}_{h\sim Q}\left[ R(h)\right] 
    &\leq \mathbb{E}_{h\sim Q}\left[ R_{m,\psi,s}(h)\right] + \frac{M_{3,s}}{s^2} +\frac{\KL(Q||P) + \log\left(\frac{1}{\delta}\right)}{m^{\alpha}}\\
    & +\frac{1}{m^{\alpha}}\log\left(\mathbb{E}_{h\sim P} \left[\exp\left( \frac{s^2}{2m^{1-2\alpha}}\psi\left(\frac{K(h)}{s}\right)^2 \right) \right]\right).
\end{align*}
\end{theorem}

\begin{proof}
The proof is similar to that of \cref{th: pac_bayes_bound_softening_functions}, we simply use \cref{l: softening functions_with strong hypothesis} instead of \cref{l: softening functions}. 
\end{proof}

\begin{remark}
    A possible choice for the pair $(\alpha,s)$ is $s^2= \sqrt{m},\alpha= 1/2$. In this way we recover the same convergence rate in $1/\sqrt{m}$ than \citet{holland2019}  for all the terms on the right-hand side of the bound. Furthermore, with those parameters we recover in the exponential moment the factor $\sqrt{m}$ also visible in $\nu^*(\mathcal{H})$ (cf \cref{hollands_theorem}). 
\end{remark}

\section{Proofs}
\label{sec:proofs}
\subsection{Proof of Theorem~\ref{th: germain2009}}
\label{proof germain2009}
\begin{proof}
    Let $D:\mathbb{R}^{+}\times\mathbb{R}^{+}\mapsto \mathbb{R}$ a convex function, $\alpha\in\mathbb{R}, P$ a fixed prior and $\delta\in[0,1]$. Since $\mathbb{E}_{h\sim P}\left[e^{m^{\alpha}D(R_m(h),R(h))} \right]$ is a nonnegative random variable, we know that, by Markov's inequality, for any $h\in\mathcal{H}$ :
    \[ 
    \mathbb{P}\left(\mathbb{E}_{h\sim P}\left[e^{m^{\alpha}D(R_m(h),R(h))} \right] > \frac{1}{\delta}\mathbb{E}_{\mathcal{S}}\hspace{1mm} \mathbb{E}_{h\sim P}\left[e^{m^{\alpha}D(R_m(h),R(h))} \right]\right) \leq \delta.     
    \]
    So with probability $1-\delta$, we have: 
    \[ 
    \mathbb{E}_{h\sim P}\left[e^{m^{\alpha}D(R_m(h),R(h))} \right] \leq \frac{1}{\delta}\mathbb{E}_{\mathcal{S}}\hspace{1mm} \mathbb{E}_{h\sim P}\left[e^{m^{\alpha}D(R_m(h),R(h))} \right].     
    \]
     
     We will now apply the logarithm function on each side of this inequality. Furthermore, because of the positiveness of $e^{m^{\alpha}D(R_m(h),R(h))}$  and because we supposed the prior $P$ to have no data dependency, we can switch the expectation symbols by Fubini-Tonelli's theorem: so with probability $1-\delta$ over samples $\mathcal{S}$, we have: 
     \[ 
     \log\left(\mathbb{E}_{h\sim P}\left[e^{m^{\alpha}D(R_m(h),R(h))} \right]\right) \leq \log\left(\frac{1}{\delta}\mathbb{E}_{h\sim P}\hspace{1mm} \mathbb{E}_{\mathcal{S}}\left[e^{m^{\alpha}D(R_m(h),R(h))} \right] \right).     
     \]
     
     We now rename $A:= \log \left(\mathbb{E}_{h\sim P}\left[e^{m^{\alpha}D(R_m(h),R(h))}  \right]\right) $.
     
     Furthermore, if we denote by $\frac{dQ}{dP}$ the Radon-Nikodym derivative of $Q$ with respect to $P$ when $Q \ll P$, we then have, for all $Q$ such that $Q \ll P$ and $P \ll Q$:
     \begin{align*}
        A 
        & = \log \left(\mathbb{E}_{h\sim Q}\left[\frac{dP}{dQ}e^{m^{\alpha}D(R_m(h),R(h))}  \right]\right)     & \\
        &= \log \left(\mathbb{E}_{h\sim Q}\left[\left(\frac{dQ}{dP}\right)^{-1}e^{m^{\alpha}D(R_m(h),R(h))}  \right]\right) 
            & \frac{dP}{dQ}= \left(\frac{dQ}{dP}\right)^{-1} \\
        & \geq -\mathbb{E}_{h\sim Q}\left[\log\left(\frac{dQ}{dP}\right)  \right] + \mathbb{E}_{h\sim Q}\left[m^{\alpha}D(R_m(h),R(h))  \right] 
            & \text{(by concavity of log with Jensen's inequality)} \\
        & \geq -\KL(Q||P) +m^{\alpha} \mathbb{E}_{h\sim Q}\left[D(R_m(h),R(h))  \right] 
            & \\
        & \geq -\KL(Q||P) + m^{\alpha}D\left(\mathbb{E}_{h\sim Q}\left[ \left(R_m(h),R(h)\right)\right]\right) 
            & \text{(by convexity of $D$ with Jensen's inequality)}\\
         & \geq -\KL(Q||P) + m^{\alpha}D\left(\mathbb{E}_{h\sim Q}\left[R_m(h)\right],\mathbb{E}_{h\sim Q}\left[R(h)\right]  \right). 
            & 
     \end{align*}
     Hence, for $Q$ such that $Q \ll P$ and $P \ll Q$,
    \begin{align*}
    D\left(\mathbb{E}_{h\sim Q}\left[ R_m(h)\right], \mathbb{E}_{h\sim Q}\left[ R(h)\right]\right) 
    &\leq \frac{1}{m^{\alpha}}\left(\KL(Q||P) + A\right).
    \end{align*}
So with probability $1-\delta$, for $Q$ such that $Q \ll P$ and $P \ll Q$,
\begin{align*}
    D\left(\mathbb{E}_{h\sim Q}\left[ R_m(h)\right], \mathbb{E}_{h\sim Q}\left[ R(h)\right]\right) 
    &\leq \frac{1}{m^{\alpha}}\left(\KL(Q||P) + \log\left(\frac{1}{\delta} \mathbb{E}_{h\sim P} \mathbb{E}_{\mathcal{S}}\; e^{m^{\alpha}D(R_m(h),R(h))}  \right)\right). 
\end{align*}
This completes the proof of Theorem~\ref{th: germain2009}.
\end{proof}

\subsection{Proof of Theorem \ref{th: pacbayes_bound_gaussian_case}}
\label{proof_gaussian_prior}
We first provide a technical property.
Recall that 
     \[ 
     \xi = \mathbb{E}_{h\sim P} \left[\exp\left(\frac{K(h)^{2}}{2m^{1-2\alpha}}\right) \right]  .
     \]
\begin{proposition}
    \label{p: xi_bound}
    Let $\alpha\in\mathbb{R}$. If the loss $\ell$ is $\SBC(K)$ compliant with $K(h)= B||h|| + C $, with $B>0$, $C\geq 0$, then we have, for any Gaussian prior $P=\mathcal{N}(0,\sigma^{2}\textbf{I}_{N})$ with $ \sigma^{2}= t\frac{m^{1-2\alpha}}{B^2}$,\hspace{1mm} $0<t<1$ and $N\geq 6$:
    \[
    \xi \leq  2\exp\left(\frac{C^2}{2m^{1-2\alpha}f(t)}\left(1+f(t)\right)\right)\frac{1}{\left(\sqrt{1-t}\right)^{N} } \left( 1+ \left(\frac{C}{\sqrt{2f(t)m^{1-2\alpha}}}\right)\right)^{N-1} 
    \]
    
    with $f(t)= \frac{1-t}{t}$.
\end{proposition}

\begin{proof}
\label{proof xi bound}
     We recall that $\sigma^2=  t\frac{m^{1-2\alpha}}{B^2}$. By expliciting the expectation and $K(h)$ we thus obtain:
     \begin{align*}
        \xi 
        & = \left( \frac{1}{\sqrt{2\pi \sigma^{2}}}\right)^{N} \int_{h\in \mathbb{R}^{N}} \exp\left(\frac{(B||h||+C)^{2}}{2m^{1-2\alpha}} - \frac{||h||^{2} B^{2}}{2tm^{1-2\alpha}} \right) dh 
            & \\
        & = \left( \frac{1}{\sqrt{2\pi \sigma^{2}}}\right)^{N} \int_{h\in \mathbb{R}^{N}} \exp\left(-\frac{1}{2m^{1-2\alpha}} \left( f(t) B^2 ||h||^{2} - 2BC||h|| - C^{2} \right) \right) dh    
            & \\
        & = \left( \frac{1}{\sqrt{2\pi \sigma^{2}}}\right)^{N} \int_{h\in \mathbb{R}^{N}} \exp\left(-\frac{B^{2}f(t)}{2m^{1-2\alpha}} \left( ||h||^{2} - \frac{2C||h||}{Bf(t)} - \frac{C^{2}}{B^2f(t)} \right) \right) dh    
            & \\
        & =  \exp\left(\frac{C^2}{2m^{1-2\alpha}f(t)}\left(1+f(t)\right)\right)\frac{1}{(\sqrt{2\pi \sigma^{2}})^{N}} \int_{h\in \mathbb{R}^{N}} \exp\left(-\frac{B^{2}f(t)}{2m^{1-2\alpha}} \left( ||h|| -\frac{C}{Bf(t)} \right)^{2} \right) dh .
     \end{align*}
     
     We will use the spherical coordinates in $N$-dimensional Euclidean space given in \cite{spherical_coord}:
     \[
     \varphi : (h_1,...,h_N)\rightarrow (r,\varphi_1,...,\varphi_{N-1})
     \]
     
     where especially $r= ||h||$ and also the Jacobian of $\phi$ is given by:
     \[
     d^NV= r^{N-1} \prod_{k=1}^{N-2} \sin^{k}(\varphi_{N-1-k}) = r^{N-1}d_{S^{N-1}}V.  
     \]
     Let us also precise that as given in \citet[][page 66]{spherical_coord}, we have that the surface of the sphere of radius 1 in $N$-dimensional space is:
     \[
     \int_{\varphi_1,...,\varphi_{N-1}}d_{S^{N-1}}V\hspace{1mm} d{\varphi_1}...d{\varphi_{N-1}}= \frac{2\sqrt{\pi}^{N}}{\Gamma\left(\frac{N}{2}\right)}   
     \]
     where $\Gamma$ is the Gamma function defined as:
     \[
     \Gamma(x)= \int_{0}^{+\infty} t^{x-1} e^{-t}dt 
     \hspace{5mm}\text{for}\hspace{2mm} x > -1.
     \]
     Then, if we set 
     \[
     A:=\int_{h\in \mathbb{R}^{N}} \exp\left(-\frac{B^{2}f(t)}{2m^{1-2\alpha}} \left( ||h|| -\frac{C}{Bf(t)} \right)^{2} \right) dh
     \]
     we obtain by a change of variable:
     \begin{align*}
         A 
         & = \int_{r,\varphi_1,...,\varphi_{N-1}}  \exp\left(-\frac{B^{2}f(t)}{2m^{1-2\alpha}} \left( r -\frac{C}{Bf(t)} \right)^{2} \right) d^{N}V\hspace{1mm} dr d{\varphi_1}...d{\varphi_{N-1}}\\
         & = \left(\frac{2\sqrt{\pi}^{N}}{\Gamma\left(\frac{N}{2}\right)}\right) \int_{r=0}^{+\infty} \exp\left(-\frac{B^{2}f(t)}{2m^{1-2\alpha}} \left( r -\frac{C}{Bf(t)} \right)^{2} \right) r^{N-1} dr \\
         & = \left(\frac{2\sqrt{\pi}^{N}}{\Gamma\left(\frac{N}{2}\right)}\right) \int_{r= -\frac{C}{Bf(t)}}^{+\infty}\left(r+\frac{C}{Bf(t)}\right)^{N-1} \exp\left(-\frac{B^{2}f(t)}{2m^{1-2\alpha}}r^{2} \right) dr \\
         & = \left(\frac{2\sqrt{\pi}^{N}}{\Gamma\left(\frac{N}{2}\right)}\right) \sum_{k=0}^{N-1} \binom{N-1}{k} \left(\frac{C}{Bf(t)}\right)^{N-k-1} \int_{r= -\frac{C}{Bf(t)}}^{+\infty}r^{k} \exp\left(-\frac{B^{2}f(t)}{2m^{1-2\alpha}}r^{2} \right) dr.
     \end{align*}
     We fix a random variable $X$ such that 
     \[
     X\sim\mathcal{N}\left(0,\frac{m^{1-2\alpha}}{B^2(f(t)}\right).
     \]
     We then have for any $k$ positive integer, if $k$ is even:
     \begin{align*}
         \int_{r= -\frac{C}{Bf(t)}}^{+\infty}r^{k} \exp\left(-\frac{B^{2}f(t)}{2m^{1-2\alpha}}r^{2} \right) dr 
         & \leq \int_{r= -\infty}^{+\infty}r^{k} \exp\left(-\frac{B^{2}f(t)}{2m^{1-2\alpha}}r^{2} \right) dr \\
         & \leq \sqrt{2\pi\frac{m^{1-2\alpha}}{B^2f(t)}} \mathbb{E}[|X|^{k}].
     \end{align*} 
     And if $k$ is odd:
     \begin{align*}
        \int_{r= -\frac{C}{Bf(t)}}^{+\infty}r^{k} \exp\left(-\frac{B^{2}f(t)}{2m^{1-2\alpha}}r^{2} \right) dr 
        & \leq \int_{r= 0}^{+\infty}r^{k} \exp\left(-\frac{B^{2}f(t)}{2m^{1-2\alpha}}r^{2} \right) dr \\
        & \leq \sqrt{2\pi\frac{ m^{1-2\alpha}}{B^2f(t)}}\mathbb{E}[|X|^{k}\mathds{1}(X\geq0)]\\
        & \leq \sqrt{2\pi\frac{ m^{1-2\alpha}}{B^2f(t)}}\mathbb{E}[|X|^{k}].
     \end{align*} 
     So we have: 
     \begin{align*}
         A 
         & \leq \left(\frac{2\sqrt{\pi}^{N}}{\Gamma\left(\frac{N}{2}\right)}\right) \sum_{k=0}^{N-1} \binom{N-1}{k} \left(\frac{C}{Bf(t)}\right)^{N-k-1} \sqrt{2\pi\frac{ m^{1-2\alpha}}{B^2f(t)}}\mathbb{E}[|X|^{k}].
     \end{align*}
     As precised in \cite{gaussian_moments}, we have for any $k$:
     \[
     \mathbb{E}[|X|^k] = \left(\sqrt{\frac{m^{1-2\alpha}}{B^2f(t)}}\right)^{k} 2^{k/2} \frac{\Gamma\left(\frac{k+1}{2}\right)}{\sqrt{\pi}}.    
     \]
     So finally:
     \begin{align*}
         A 
         &\leq 2\sqrt{\pi}^{N}\hspace{1mm} \sum_{k=0}^{N-1} \binom{N-1}{k} \left(\frac{C}{Bf(t)}\right)^{N-k-1} \left(\sqrt{\frac{2m^{1-2\alpha}}{B^2f(t)}}\right)^{k+1} \frac{\Gamma\left(\frac{k+1}{2}\right)}{\Gamma\left(\frac{N}{2} \right)}. 
     \end{align*}
     \begin{lemma}
        \label{l: gamma}
        If $N\geq 6$, then:
        \[
        \max_{k=0..N-1} \frac{\Gamma\left(\frac{k+1}{2}\right)}{\Gamma\left(\frac{N}{2}\right)} = 1   .
        \]
     \end{lemma}
     
     \begin{proof}
     \noindent
          As precised in the introduction of \citet{srinivasan_2011}, \citet[][page 147]{gauss_2011}\footnote{Do not mind the year in that reference: this is obviously a reprint!} proved that on the interval $ [x_0,+\infty)$ where $x_0\in[1.46,1.47]$, $\Gamma$ is a monotonic increasing function. So, for $N-1\geq k\geq 2, \Gamma(\frac{k+1}{2})\leq \Gamma(\frac{N}{2})$.
          And because $\Gamma(1/2)=\sqrt{\pi},\Gamma(1)=1$, we have
          \[ 
          \max_{k=0..N-1} \frac{\Gamma\left(\frac{k+1}{2}\right)}{\Gamma\left(\frac{N}{2}\right)} = \max\left(\frac{\sqrt{\pi}}{\Gamma\left(\frac{N}{2}\right)}, \frac{\Gamma\left(\frac{N-1+1}{2}\right)}{\Gamma\left(\frac{N}{2}\right)}\right) = \max \left(\frac{\sqrt{\pi}}{\Gamma\left(\frac{N}{2}\right)}, 1\right)   
          \] 
          \noindent
          And because $N\geq 6$ and that $\Gamma$ is monotone increasing on $[3;+\infty]$, we have $\Gamma(N/2)\geq \Gamma(3)\geq \sqrt{\pi}$. Hence the result.
     \end{proof}
     Using \cref{l: gamma} allows us to write:
     \begin{align*}
         A 
         &\leq 2\sqrt{\pi}^{N}\hspace{1mm} \sum_{k=0}^{N-1} \binom{N-1}{k} \left(\frac{C}{Bf(t)}\right)^{N-k-1} \left(\sqrt{\frac{2m^{1-2\alpha}}{B^2f(t)}}\right)^{k+1}. 
     \end{align*}
     We recall that $ \sigma^2 = t\frac{m^{1-2\alpha}}{B^2}$ and $f(t)=\frac{1-t}{t}$. Then we can write: 
     \[
     A\leq 2\sqrt{\pi}^{N}\hspace{1mm} \sum_{k=0}^{N-1} \binom{N-1}{k} \left(\frac{C}{Bf(t)}\right)^{N-k-1} \left(\sqrt{\frac{2\sigma^{2}}{1-t}}\right)^{k+1}  .
     \]
     We now conclude with the final bound on $\xi$
     \begin{align*}
        \xi 
        & \leq  \exp\left(\frac{C^2}{2m^{1-2\alpha}f(t)}\left(1+f(t)\right)\right)\frac{1}{(\sqrt{2\pi \sigma^{2}})^{N}} \hspace{1mm} A \\
        & \leq \exp\left(\frac{C^2}{2m^{1-2\alpha}f(t)}\left(1+f(t)\right)\right)\frac{1}{(\sqrt{2\pi \sigma^{2}})^{N}} 2\sqrt{\pi}^{N}\hspace{1mm} \sum_{k=0}^{N-1} \binom{N-1}{k} \left(\frac{C}{Bf(t)}\right)^{N-k-1} \left(\sqrt{\frac{2\sigma^{2}}{1-t}}\right)^{k+1} \\
        & \leq 2\exp\left(\frac{C^2}{2m^{1-2\alpha}f(t)}\left(1+f(t)\right)\right) \sum_{k=0}^{N-1} \binom{N-1}{k} \left(\frac{C}{Bf(t)}\right)^{N-k-1} \left(\sqrt{\frac{1}{1-t}}\right)^{k+1} \left(\sqrt{\frac{B^2}{2tm^{1-2\alpha}}}\right)^{N-k-1}\\
        & \leq 2\exp\left(\frac{C^2}{2m^{1-2\alpha}f(t)}\left(1+f(t)\right)\right) \sum_{k=0}^{N-1} \binom{N-1}{k} \left(\frac{C\sqrt{t}}{(1-t)\sqrt{2m^{1-2\alpha}}}\right)^{N-k-1} \left(\sqrt{\frac{1}{1-t}}\right)^{k+1}\\
        & \leq 2\frac{\exp\left(\frac{C^2}{2m^{1-2\alpha}f(t)}\left(1+f(t)\right)\right)}{\left(\sqrt{1-t}\right)^{N} } \sum_{k=0}^{N-1} \binom{N-1}{k} \left(\frac{C}{\sqrt{2f(t)m^{1-2\alpha}}}\right)^{N-k-1}\\
        & \leq 2\frac{\exp\left(\frac{C^2}{2m^{1-2\alpha}f(t)}\left(1+f(t)\right)\right)}{\left(\sqrt{1-t}\right)^{N} } \left( 1+ \left(\frac{C}{\sqrt{2f(t)m^{1-2\alpha}}}\right)\right)^{N-1}.
     \end{align*}
This completes the proof of \cref{p: xi_bound}.
\end{proof}

\begin{proof}[of \cref{th: pacbayes_bound_gaussian_case}].
We just have to articulate \cref{th: pac_bayes_with_self_bounding} and \cref{p: xi_bound} altogether. We also upper-bound $N-1$ by $N$.
\end{proof}

\end{document}